\documentclass[sigconf,twocolumn]{acmart}

\usepackage{booktabs} % For formal tables
\usepackage{cleveref}
\usepackage{multirow}
\usepackage{tikz,pgfplots}

\setcopyright{rightsretained}
\acmDOI{10.1145/nnnnnnn.nnnnnnn}

\acmISBN{978-x-xxxx-xxxx-x/YY/MM}

\acmConference[GECCO '20]{the Genetic and Evolutionary Computation Conference 2020}{July 8--12, 2020}{Cancun, Mexico}
\acmYear{2020}
\copyrightyear{2020}

\acmPrice{15.00}

\newcommand{\A}{\mbox{$\mathsf{A}$}}
\newcommand{\B}{\mbox{$\mathsf{B}$}}
\newcommand{\D}{\mbox{$\mathsf{D}$}}

\newcommand{\Largest}{\mbox{$\mathsf{L}$}}
\newcommand{\Smallest}{\mbox{$\mathsf{S}$}}
\newcommand{\MS}{\mbox{$\mathsf{MS}$}}
\newcommand{\CA}{\mbox{$\mathsf{CA}$}}
\newcommand{\CAN}{\mbox{$\mathsf{CAN}$}}

\newcommand{\MCA}{\mbox{$\mathsf{MCA}$}}

\begin{document}
	\title{Genetic Algorithms for Redundancy in Interaction Testing}
	
	\author{Ryan E. Dougherty}
	\orcid{0000-0003-1739-1127}
	\affiliation{%
		\institution{Colgate University}
		\streetaddress{13 Oak Drive}
		\city{Hamilton} 
		\state{New York} 
		\postcode{13346}
	}
	\email{rdougherty@colgate.edu}
	
	% The default list of authors is too long for headers.
	\renewcommand{\shortauthors}{R. E. Dougherty}

	\begin{abstract}
		It is imperative for testing to determine if the components within large-scale software systems operate functionally.
		Interaction testing involves designing a suite of tests, which guarantees to detect a fault if one exists among a small number of components interacting together.
		The cost of this testing is typically modeled by the number of tests, and thus much effort has been taken in reducing this number.
		Here, we incorporate redundancy into the model, which allows for testing in non-deterministic environments.
		Existing algorithms for constructing these test suites usually involve one ``fast'' algorithm for generating most of the tests, and another ``slower'' algorithm to ``complete'' the test suite.
		We employ a genetic algorithm that generalizes these approaches that also incorporates redundancy by increasing the number of algorithms chosen, which we call ``stages.''
		By increasing the number of stages, we show that not only can the number of tests be reduced compared to existing techniques, but the computational time in generating them is also greatly reduced.
	\end{abstract}
	
	%
	% The code below should be generated by the tool at
	% http://dl.acm.org/ccs.cfm
	% Please copy and paste the code instead of the example below. 
	%
	\begin{CCSXML}
		<ccs2012>
		<concept>
		<concept_id>10011007.10011074.10011784</concept_id>
		<concept_desc>Software and its engineering~Search-based software engineering</concept_desc>
		<concept_significance>500</concept_significance>
		</concept>
		</ccs2012>
	\end{CCSXML}
	
	\ccsdesc[500]{Software and its engineering~Search-based software engineering}
	
	\keywords{covering array, genetic algorithm, redundancy, search-based software engineering}

	\maketitle
	
	\section{Introduction}
	
	The task of testing software systems has always remained a challenging problem, in that a tester is to find and eliminate any defects that were not found during earlier stages of development.
	One systematic approach has been interaction testing 
	\cite{kuhn2004software,dalal1999model,yilmaz2006covering}, 
	wherein a system configuration is presented, and the task is to design a series of tests such that any interaction between components at most a certain size is tested.
	It is empirically shown in \cite{kuhn2002investigation} that a very large percentage of errors within a software system can be detected with interactions of small size.
	
	Suppose that a system to be tested has \textit{factors} $f_1, \cdots, f_k$, and each factor $f_i$ has a set of allowed \textit{levels} (i.e., valid inputs to $f_i$).
	Let $t$ be a positive integer at most $k$, the number of factors.
	We design a test suite (i.e., a set of tests) such that each one is a valuation of each factor to one of its valid inputs, with the property that any set of $t$ or fewer factors is exhaustively tested.
	By this last term, we mean that no matter the set of $t$ factors chosen, and any level selected for each factor, at least one test has that choice.
	The parameter $t$ is the \textit{strength} of the test suite.
	
	We present an equivalent formulation.
	Define an \textit{interaction of size $t$} (or a \textit{$t$-way interaction}) to be a set of the form
	\[
	\{(f_i, \ell_i) : \ell_i\;\text{is a level for factor $f_i$}, 1 \le i \le t\}.
	\]
	Let $\mathcal{I}_{k,t}$ be the set of all interactions of size at most $t$ with $k$ factors.
	Then we say an array $A$ with $N$ rows and $k$ columns is a \textit{mixed-level covering array} $\MCA(N; t, (v_1, \cdots, v_k))$ if (1) column $i$ corresponds to factor $f_i$, (2) $v_i$ is the number of levels for $f_i$, and (3) all interactions in $\mathcal{I}_{k,t}$ appear at least once in the array.
	When $v_1 = \cdots = v_k = v$, we say that the array $A$ is a (uniform) \textit{covering array} $\CA(N; t, k, v)$. 
	This paper only concerns uniform covering arrays.
	
	\begin{figure}
		\centering
		
		\begin{tabular}{|c|c|c|c|}
			\hline
			Browser &  OS &  Connection &  Printer \\
			\hline 
			Safari &  Windows &  LAN &  Local \\
			Safari &  Linux &  ISDN &  Networked \\
			Safari &  macOS &  PPP &  Screen \\
			IE &  Windows &  ISDN &  Screen \\
			IE &  macOS &  LAN &  Networked \\
			IE &  Linux &  PPP &  Local \\
			Chrome &  Windows &  PPP &  Networked \\
			Chrome &  Linux &  LAN &  Screen \\
			Chrome &  macOS &  ISDN &  Local \\
			\hline
		\end{tabular}
		\caption{A Covering Array with 9 tests ($N=9$), 4 factors ($k=4$), each with 3 levels ($v=3$), and strength 2 ($t=2$). \label{tbl:test_suite}}
		
	\end{figure}
	
	To test a system using a covering array, a tester observes each test in turn, and chooses the corresponding level for each factor as dictated by the test.
	Observe the test suite in \Cref{tbl:test_suite}, reproduced from \cite{colbourn2004combinatorial}; we claim it is a $\CA(9;2,4,3)$.
	There are 9 rows (representing the tests), 4 factors (Browser, OS, Connection, and Printer), 3 levels for each factor (e.g., Safari, IE, and Chrome for the factor Browser), and has strength 2.
	The tester then records whether the output of the system, after the test is performed, is expected; if this property is true for all tests, then the coverage property guarantees no fault exists in the system that involves at most $t$ factors.
	If a fault exists within this system, it must be due to an interaction of 3 or more factors, since all 2-way interactions appear at least once.
	With covering arrays we develop in this paper, we substitute the levels in each factor for 0, 1, 2, $\cdots$ for ease of presentation.
	
	With the standard definition, we assume that the system is \textit{deterministic}, in that running a given test multiple times always produces the same result.
	For example, in the $\CA$ from \Cref{tbl:test_suite}, the first test of Safari/Windows/LAN/Local, if repeated, would produce the same output.
	% In most uses of interaction testing, one assumes that the system behaves deterministically, i.e., when a test is run multiple times, the same output occurs.
	However, the assumption of the system's being deterministic is often not realistic, particularly when there is \textit{noise} or \textit{randomness} within the testing environment.
	If we amend our definition to say that each interaction appears \textit{at least a given number of times $\lambda$}, then by increasing $\lambda$, we gain further confidence in the correctness of the system, even if it is not deterministic.
	For example, if $\lambda = 5$, then if the resulting output matches what was expected 5 times in a row, it is much less likely that a fault still exists than if, say, $\lambda = 1$.
	
	We update our notation with \textit{index $\lambda$} as follows: $\CA_\lambda(N; t, k, v)$, in that every interaction appears at least $\lambda$ times.
	It is possible to generalize our model by having each interaction $I$ have its own corresponding index $\lambda_I$, but we content ourselves with the generality developed here, since we can take the maximum index over all interactions to be the $\lambda$ we seek.
	
	To minimize total cost of the tester, the most often chosen metric for a covering array of ``smallest cost''  is finding the smallest number of tests $N$ for which such an array exists.
	The \textit{covering array number}, $\CAN_\lambda(t, k, v)$, is this quantity.
	Much work has been to determine these values for $\lambda = 1$ \cite{colbourn2004combinatorial,sarkar2017upper,colbourn2011covarrayandhashfamily}, but none for when $\lambda > 1$, as far as we are aware.
	When $\lambda > 1$, we say that the array is of \textit{higher-index}.
	
	Sarkar and Colbourn \cite{sarkar_colbourn_twostage} introduced the \textit{two-stage} framework for constructing covering arrays, which is as follows.
	The first stage generates an array randomly which covers all but (at most) a certain number of interactions, and the second stage deterministically covers the remaining interactions with more rows until all interactions are covered.
	Their methods determine the smallest number of rows in the first stage such that the expected number of rows produced after the second stage is minimized. 
	They determined that in the second stage there is a trade-off between computation time and bound on $\CAN_\lambda$, and no one method is uniformly better than the others.
	Further, the first stage is comparatively faster than the second, but suffers from not giving a guarantee on the exact number of interactions left for the second stage, or their structure.
	And finally, they only studied the $\lambda=1$ case.
	
	There have been much previous work involving genetic algorithms and other metaheuristical techniques for covering arrays and related objects \cite{nurmela2004upper,stardom2001metaheuristics,leach19,timana2016metaheuristic,dougherty19gecco}.
	However, our approach is different because our genetic algorithm is focused on determining an optimal number of \textit{constructive} stages (as well as which method to use for each), whereas all existing techniques focus on mutations to the array itself in the hope of forming a $\CA$.
	An main advantage of our approach is that the algorithms chosen are fully deterministic and replicable.
	
	The contributions of this paper are that
	\begin{enumerate}
		\item we extend the methods of Sarkar and Colbourn by introducing a \textit{multi-stage} framework for higher-index covering arrays using a genetic algorithm;
		\item our framework is general enough to allow for any choice of algorithm at any stage, with any number of stages, and any index;
		\item our genetic algorithm's choices of which stages to select, and how many of them, yield a dramatic decrease in computational time in the creation of these arrays, sometimes by two orders of magnitude; and
		\item we show that there is a pattern to which algorithms are best to choose at each stage by analyzing the Pareto fronts resulting from our genetic algorithm.
	\end{enumerate}
	
	\section{Our Multi-Stage Frameowrk}
	
	Denote $\MS\langle \A_1(\lambda_1), \cdots, \A_M(\lambda_m)\rangle$ to be a \textit{multi-stage selection with algorithms $\A_1, \cdots, \A_m$}, such that after algorithms $\A_1, \cdots, \A_i$ are applied, then a covering array is produced with index $\lambda_1 + \cdots + \lambda_i$.
	If the desired index is $\lambda$, then it is the case that $\lambda = \lambda_1 + \cdots + \lambda_m$.
	We call the application of $\A_i$ to be the \textit{$i$th stage}.
	Once the $i$th stage is completed, the resulting array is fed as input to the $(i+1)$-st stage.
	The goal of each algorithm $\A_i$ is to be computationally efficient, require few rows to complete, and cover many interactions to reduce the cost of future stages. 

	\subsection{Why Multiple Stages?}
	
	It is not immediately clear why multiple stages may be advantageous over a single stage. 
	For multiple stages, a software tester would have to choose which algorithms to use in each stage, what order to choose them, what index to choose for each, and above all, there is not necessarily any communication between the stages on how to optimize the parameters for each.
	A single stage also is easier to implement and maintain, often has performance guarantees (in terms of run time to produce the covering array and its final size), and the produced arrays often have a structure that can be easily analyzed, since the method to construct it is fixed.
	
	We give a simple, but representative example of why multiple stages make sense, and an intuitive understanding why; a more detailed explanation is in \Cref{sec:discussion}.
	The Unix {\tt sort} command has 18 binary flags; an exhaustive test suite to test the correctness of {\tt sort} would involve $2^{18}$ tests!
	Suppose that we want to build a test suite with $t=2$; the best known (and proven optimal) covering array with $\lambda = 1$ has 8 rows \cite{kleitman1973families}.
	Now suppose we desire to have a redundancy of $\lambda = 5$; in other words, we desire to build a $\CA_5(N; 2, 18, 2)$, with $N$ as small as possible.
	A na\"ive approach would say that 40 rows is possible, simply by juxtaposing the 8-row array 5 times.
	
	An implementation of an extension to a well-known greedy algorithm (described in \Cref{subsec:density}) produces an array with 29 rows in a single stage.
	However, a simple choice of algorithm selection found that there exist 3-stage and 4-stage solutions yielding 27 rows; furthermore, all of these multiple-stage solutions complete in 50\% or less time compared to the single-stage one.
	In fact, all solutions found between 27 and 28 rows had 2 or more stages, among all algorithms tested.
	{\setlength{\tabcolsep}{0pt}
		
		\begin{figure}[]
			\begin{tabular}{llllllllllllllllll}
				0 & 0 & 0 & 0 & 0 & 0 & 0 & 0 & 0 & 0 & 0 & 0 & 0 & 0 & 0 & 0 & 0 & 0\\
				1 & 1 & 1 & 1 & 1 & 1 & 1 & 1 & 1 & 1 & 1 & 1 & 1 & 1 & 1 & 1 & 1 & 0\\
				0 & 1 & 0 & 1 & 0 & 1 & 0 & 1 & 0 & 1 & 0 & 1 & 0 & 1 & 0 & 1 & 0 & 1\\
				1 & 0 & 1 & 0 & 1 & 0 & 1 & 0 & 1 & 0 & 1 & 0 & 1 & 0 & 1 & 0 & 1 & 1\\
				0 & 0 & 1 & 1 & 0 & 0 & 1 & 1 & 0 & 0 & 1 & 1 & 0 & 0 & 1 & 1 & 0 & 0\\
				1 & 1 & 0 & 0 & 1 & 1 & 0 & 0 & 1 & 1 & 0 & 0 & 1 & 1 & 0 & 0 & 0 & 0\\
				0 & 0 & 0 & 0 & 1 & 1 & 1 & 1 & 0 & 0 & 0 & 0 & 1 & 1 & 1 & 1 & 1 & 0\\
				1 & 1 & 1 & 1 & 0 & 0 & 0 & 0 & 1 & 1 & 1 & 1 & 0 & 0 & 0 & 0 & 1 & 0\\
				0 & 0 & 0 & 0 & 0 & 0 & 0 & 0 & 1 & 1 & 1 & 1 & 1 & 1 & 1 & 1 & 0 & 0\\
				1 & 1 & 1 & 1 & 1 & 1 & 1 & 1 & 0 & 0 & 0 & 0 & 0 & 0 & 0 & 0 & 0 & 0\\
				\hline 
				1 & 0 & 0 & 1 & 0 & 1 & 1 & 0 & 0 & 1 & 1 & 0 & 1 & 0 & 0 & 1 & 0 & 1\\
				0 & 1 & 1 & 0 & 1 & 0 & 0 & 1 & 1 & 0 & 0 & 1 & 0 & 1 & 1 & 0 & 1 & 1\\
				0 & 0 & 0 & 0 & 0 & 0 & 0 & 0 & 0 & 0 & 0 & 0 & 0 & 0 & 0 & 0 & 1 & 0\\
				\hline
				1 & 1 & 1 & 0 & 0 & 1 & 1 & 1 & 1 & 1 & 0 & 1 & 1 & 0 & 1 & 1 & 1 & 1\\
				0 & 0 & 1 & 1 & 1 & 0 & 0 & 0 & 0 & 1 & 1 & 0 & 1 & 1 & 0 & 1 & 1 & 1\\
				1 & 1 & 0 & 1 & 1 & 0 & 1 & 1 & 1 & 0 & 1 & 1 & 0 & 1 & 0 & 0 & 0 & 1\\
				0 & 1 & 0 & 1 & 0 & 1 & 0 & 0 & 1 & 0 & 1 & 0 & 0 & 1 & 1 & 0 & 1 & 1\\
				1 & 0 & 1 & 0 & 1 & 1 & 0 & 1 & 0 & 1 & 1 & 1 & 0 & 0 & 1 & 0 & 0 & 1\\
				0 & 1 & 1 & 1 & 0 & 0 & 1 & 0 & 0 & 0 & 0 & 1 & 1 & 1 & 0 & 1 & 0 & 1\\
				1 & 0 & 0 & 0 & 1 & 1 & 1 & 1 & 1 & 1 & 0 & 0 & 0 & 0 & 1 & 1 & 0 & 0\\
				0 & 1 & 0 & 0 & 1 & 1 & 1 & 0 & 0 & 0 & 1 & 1 & 1 & 0 & 1 & 0 & 1 & 0\\
				1 & 0 & 1 & 1 & 0 & 0 & 0 & 1 & 1 & 0 & 0 & 0 & 1 & 1 & 0 & 1 & 1 & 0\\
				0 & 0 & 1 & 0 & 0 & 1 & 1 & 0 & 1 & 1 & 0 & 1 & 0 & 1 & 0 & 0 & 1 & 1\\
				1 & 1 & 0 & 1 & 1 & 0 & 0 & 1 & 0 & 1 & 1 & 0 & 1 & 0 & 1 & 0 & 0 & 1\\
				0 & 1 & 0 & 0 & 1 & 0 & 1 & 0 & 1 & 1 & 1 & 0 & 0 & 0 & 0 & 1 & 0 & 0\\
				1 & 0 & 0 & 1 & 0 & 1 & 0 & 0 & 0 & 0 & 0 & 1 & 0 & 1 & 1 & 1 & 1 & 0\\
				0 & 0 & 1 & 0 & 0 & 0 & 0 & 1 & 0 & 0 & 0 & 0 & 1 & 0 & 0 & 0 & 0 & 0\\
				&&&&&&&&&&&&&&&&&\\
				&&&&&&&&&&&&&&&&&
			\end{tabular}
			\quad
			\begin{tabular}{llllllllllllllllll}
				0 & 0 & 0 & 0 & 0 & 0 & 0 & 0 & 0 & 0 & 0 & 0 & 0 & 0 & 0 & 0 & 0 & 0 \\
				1 & 1 & 1 & 1 & 1 & 1 & 1 & 1 & 1 & 1 & 1 & 1 & 1 & 1 & 1 & 1 & 1 & 0\\
				0 & 1 & 0 & 1 & 0 & 1 & 0 & 1 & 0 & 1 & 0 & 1 & 0 & 1 & 0 & 1 & 0 & 1\\
				1 & 0 & 1 & 0 & 1 & 0 & 1 & 0 & 1 & 0 & 1 & 0 & 1 & 0 & 1 & 0 & 1 & 1\\
				0 & 0 & 1 & 1 & 0 & 0 & 1 & 1 & 0 & 0 & 1 & 1 & 0 & 0 & 1 & 1 & 0 & 0\\
				1 & 1 & 0 & 0 & 1 & 1 & 0 & 0 & 1 & 1 & 0 & 0 & 1 & 1 & 0 & 0 & 1 & 0\\
				0 & 1 & 1 & 0 & 1 & 0 & 0 & 1 & 0 & 1 & 1 & 0 & 1 & 0 & 0 & 1 & 0 & 1\\
				1 & 0 & 0 & 1 & 0 & 1 & 1 & 0 & 1 & 0 & 0 & 1 & 0 & 1 & 1 & 0 & 1 & 1\\
				0 & 0 & 0 & 0 & 1 & 1 & 1 & 1 & 1 & 1 & 1 & 1 & 0 & 0 & 0 & 0 & 0 & 0\\
				1 & 1 & 1 & 1 & 0 & 0 & 0 & 0 & 0 & 0 & 0 & 0 & 1 & 1 & 1 & 1 & 1 & 0\\
				0 & 1 & 0 & 1 & 1 & 0 & 1 & 0 & 0 & 1 & 0 & 1 & 1 & 0 & 1 & 0 & 1 & 1\\
				1 & 0 & 1 & 0 & 0 & 1 & 0 & 1 & 1 & 0 & 1 & 0 & 0 & 1 & 0 & 1 & 0 & 1\\
				0 & 0 & 1 & 1 & 1 & 1 & 0 & 0 & 0 & 0 & 1 & 1 & 1 & 1 & 0 & 0 & 0 & 0\\
				1 & 1 & 0 & 0 & 0 & 0 & 1 & 1 & 1 & 1 & 0 & 0 & 0 & 0 & 1 & 1 & 1 & 0\\
				0 & 1 & 1 & 0 & 0 & 1 & 1 & 0 & 1 & 0 & 0 & 1 & 1 & 0 & 0 & 1 & 1 & 1\\
				1 & 0 & 0 & 1 & 1 & 0 & 0 & 1 & 0 & 1 & 1 & 0 & 0 & 1 & 1 & 0 & 0 & 1\\
				0 & 0 & 0 & 0 & 0 & 0 & 0 & 0 & 1 & 1 & 1 & 1 & 1 & 1 & 1 & 1 & 0 & 0\\
				1 & 1 & 1 & 1 & 1 & 1 & 1 & 1 & 0 & 0 & 0 & 0 & 0 & 0 & 0 & 0 & 1 & 0\\
				0 & 1 & 0 & 1 & 0 & 1 & 0 & 1 & 1 & 0 & 1 & 0 & 1 & 0 & 1 & 0 & 1 & 1\\
				1 & 0 & 1 & 0 & 1 & 0 & 1 & 0 & 0 & 1 & 0 & 1 & 0 & 1 & 0 & 1 & 0 & 1\\
				0 & 0 & 1 & 1 & 0 & 0 & 1 & 1 & 1 & 1 & 0 & 0 & 1 & 1 & 0 & 0 & 1 & 0\\
				1 & 1 & 0 & 0 & 1 & 1 & 0 & 0 & 0 & 0 & 1 & 1 & 0 & 0 & 1 & 1 & 0 & 0\\
				0 & 0 & 0 & 0 & 1 & 1 & 1 & 1 & 0 & 0 & 0 & 0 & 1 & 1 & 1 & 1 & 0 & 0\\
				1 & 1 & 1 & 1 & 0 & 0 & 0 & 0 & 1 & 1 & 1 & 1 & 0 & 0 & 0 & 0 & 1 & 0\\
				0 & 0 & 0 & 0 & 0 & 0 & 0 & 0 & 0 & 0 & 0 & 0 & 0 & 0 & 0 & 0 & 1 & 0\\
				1 & 1 & 1 & 1 & 1 & 1 & 1 & 1 & 1 & 1 & 1 & 1 & 1 & 1 & 1 & 1 & 0 & 0\\
				0 & 0 & 0 & 0 & 0 & 0 & 0 & 0 & 0 & 0 & 0 & 0 & 0 & 0 & 0 & 0 & 0 & 0\\
				0 & 0 & 0 & 0 & 0 & 0 & 0 & 0 & 0 & 0 & 1 & 0 & 0 & 0 & 0 & 1 & 1 & 0\\
				0 & 1 & 0 & 0 & 0 & 0 & 0 & 0 & 1 & 0 & 0 & 0 & 0 & 0 & 0 & 0 & 0 & 0 
			\end{tabular}
			
			\caption{A $\CA_5(27;2,18,2)$ (left) and a $\CA_5(29; 2, 18, 2)$ (right).\label{fig:two_example_cas}}
		\end{figure}

		\begin{figure}[]
			
		\end{figure}
	}
	
	An intuitive understanding of why multiple stages are better is that the algorithm produces the array one row at a time, and is ``greedy'' in the sense that many interactions are covered early, and only a small number are left uncovered in the last rows.
	Consider the two $\CA_5$s in \Cref{fig:two_example_cas}; the one with $N=29$ rows (right) was produced via this greedy method, and the one with $N=27$ rows (left) was produced with 3 stages, the first and third being this same greedy method, and the second via graph coloring.
	For the latter, the first and second stage has an index of 1, and the third stage has an index of 3.
	Therefore, the selection was $\MS\langle G(1), C(1), G(1) \rangle$, where $G$ represents greedy, and $C$ represents coloring.
	The horizontal lines in the left $\CA_3$ indicate where one stage ends, and another begins.
	Not only is there a reduction in the rows, but the left $\CA$ was produced in 75\% less time than the right $\CA$.
	
\begin{figure}
		\begin{tikzpicture}
		\begin{axis}[%
		scatter/classes={%
			a={mark=o,draw=black},
			b={mark=x,draw=blue}},
		xlabel={Row Number},
		ylabel={\# Interactions $\lambda$-covered},
		]
		\addplot[scatter,only marks,%
		scatter src=explicit symbolic]%
		table[meta=label] {
			x y label
			0 0 b
			1 0 b
			2 0 b
			3 0 b
			4 0 b
			5 0 b
			6 0 b
			7 0 b
			8 1 b
			9 2 b
			10 21 b
			11 40 b
			12 60 b
			13 80 b
			14 80 b
			15 80 b
			16 193 b
			17 306 b
			18 373 b
			19 440 b
			20 476 b
			21 512 b
			22 546 b
			23 580 b
			24 592 b
			25 604 b
			26 606 b
			27 610 b
			28 612 b
			0 0 a
			1 0 a
			2 0 a
			3 0 a
			4 0 a
			5 0 a
			6 0 a
			7 0 a
			8 0 a
			9 1 a
			10 5 a
			11 5 a
			12 101 a
			13 159 a
			14 194 a
			15 231 a
			16 267 a
			17 309 a
			18 355 a
			19 402 a
			20 445 a
			21 490 a
			22 529 a
			23 565 a
			24 585 a
			25 606 a
			26 612 a
		};
		\end{axis}
		
		\end{tikzpicture}
		\caption{Number of interactions first $\lambda$-covered in each row of the $\CA_5(27; 2, 18, 2)$ (circles) and $\CA_5(29; 2, 18, 2)$ (crosses) in \Cref{fig:two_example_cas}.\label{plot_private_interaction_example}}
	\end{figure}
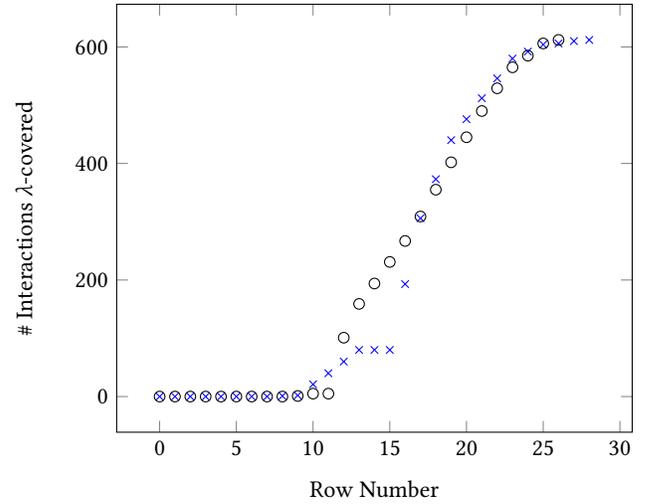

	We investigate these two $\CA$s further.
	In \Cref{plot_private_interaction_example}, we give a scatter plot corresponding to each of the $\CA$s as follows.
	The horizontal axis indicates row numbers, and for a particular row $i$, we calculate the number of interactions $N_i$ such that they become $\lambda$-covered in row $i$.
	For example, for the left $\CA$ in \Cref{fig:two_example_cas}, in the first two columns, the two values $00$ appear in rows 0, 4, 6, 8, and 12, among others. 
	Therefore, we have that the interaction $\{(0,0), (1,0)\}$ becomes 5-covered in row 12.
	The vertical axis in the plot indicates a cumulative total of interactions that are 5-covered.
	The first CA's plot achieves a vertical value of ${k \choose t}v^t = 612$ at $N = 27$, and the second at $N = 29$.
	We can see that when the first stage ends after row 10, the plot for the left $\CA$ increases more quickly than for the right $\CA$.
	What is less clear is why, even though the right $\CA$ overtakes the left, it still requires more rows.
	
	A simple analysis of the interactions left to be covered in the final rows (for the right $\CA$) shows that there are many \textit{conflicts} between the interactions that force them to appear in separate rows.
	However, the algorithm used for this $\CA$ inherently is a one-row-at-a-time method and cannot ``backtrack'' to make any necessary changes to earlier rows.
	Even though the left $\CA$ uses the same algorithm, it has multiple stages, and thus the stages can ``communicate'' information about the remaining interactions, whereas the right $\CA$ cannot since it uses only one stage.
	
	We ran a systematic experiment to further illustrate why multiple stages are advantageous.
	Our experiment involved all algorithms explained in \Cref{sec:algorithms}, and all decompositions of a target index $\lambda=5$ into at most 5 choices of algorithms (possibly repeating).
	We were able to speed up the computation substantially by noting that all choices of algorithms described in \Cref{sec:algorithms} are deterministic.
	Therefore, for any two multi-stage strategies starting with the same algorithm and index can save the result of the first stage without having to repeat the computation ourselves. 
	We saved the time it took to generate that initial array, and added its creation time to the total computation time when the final array is constructed.
	
	The results of this experiment appear in \Cref{tbl:statistics_of_example_method}.
	One can easily see that the number of rows varies substantially depending on the number of stages taken.
	As the number of stages increases, the maximum, average, and standard deviation for the number of produced rows all decrease.
	If one were to pick a number of stages a priori and then choose a set of methods uniformly at random, the obvious best choice would be 5 stages in this case, since the average is the smallest.
	As an added benefit, the minimum, maximum, average, and median time taken to construct these arrays also decrease, some by over an order of magnitude.
	
	\begin{table*}[t]
		\begin{tabular}{|l|l|l|l|l|l|l|l|l|l|l|}
			\hline
			$NS$ & Min $N$ & Max $N$ & Avg $N$ & Median $N$ & StdDev $N$ & Min $T$ & Max $T$ & Avg $T$ & Median $T$ & StdDev $T$ \\ \hline
			1   &  29   &  3060   &  826 & 138 &  1481 & 0.361 & 17.129 & 6.943 & 4.870 & 6.460  \\ \hline
			2   &  27   &  2448   &  541 & 119  & 738 & 0.014 & 11.614 & 2.692 & 1.785 & 3.141 \\ \hline
			3   &  27  & 1836 &  388 & 130  & 473 & 0.013 & 7.076 & 1.318 & 0.802 & 1.623   \\ \hline
			4   &  27   &  1224   & 297 & 134 & 323 & 0.013 & 4.006 & 0.742 & 0.393 & 0.886   \\ \hline
			5   &  28   &  627   & 239 & 130 & 228 & 0.013 & 1.815 & 0.455 & 0.264 & 0.457    \\ \hline
		\end{tabular}
		\caption{Statistics of the number of stages $NS$ when $t=2,k=18,v=2,\lambda=5$, where $N$ is the number of rows produced, and $T$ is the total computational time (in seconds). 
			The minimum, maximum, average, median, and standard deviation for both $N$ and $T$ are given.
			The ``Avg $N$,'' ``Med $N$,'' and ``StdDev $N$'' columns are rounded to the nearest integer.
			\label{tbl:statistics_of_example_method}}
	\end{table*}
	
	% \begin{figure}
	% \begin{tikzpicture}
	% \begin{axis}[%
	% scatter/classes={%
	%     a={mark=o,draw=black},
	%     b={mark=x,draw=blue}}]
	% \addplot[scatter,only marks,%
	%     scatter src=explicit symbolic]%
	% table[meta=label] {
	% x y label
	% 0 153 a
	% 1 306 a
	% 2 459 a
	% 3 612 a
	% 4 765 a
	% 5 918 a
	% 6 1071 a
	% 7 1224 a
	% 8 1377 a
	% 9 1530 a
	% 10 1683 a
	% 11 1836 a
	% 12 1989 a
	% 13 2140 a
	% 14 2272 a
	% 15 2401 a
	% 16 2514 a
	% 17 2623 a
	% 18 2715 a
	% 19 2796 a
	% 20 2861 a
	% 21 2927 a
	% 22 2971 a
	% 23 3012 a
	% 24 3033 a
	% 25 3054 a
	% 26 3060 a
	% 0 153 b
	% 1 306 b
	% 2 459 b
	% 3 612 b
	% 4 765 b
	% 5 918 b
	% 6 1071 b
	% 7 1224 b
	% 8 1377 b
	% 9 1530 b
	% 10 1683 b
	% 11 1836 b
	% 12 1979 b
	% 13 2122 b
	% 14 2245 b
	% 15 2368 b
	% 16 2509 b
	% 17 2650 b
	% 18 2745 b
	% 19 2840 b
	% 20 2895 b
	% 21 2950 b
	% 22 2987 b
	% 23 3024 b
	% 24 3038 b
	% 25 3052 b
	% 26 3054 b
	% 27 3058 b
	% 28 3060 b
	
	%     };
	% \end{axis}
	% \end{tikzpicture}
	% \end{figure}

	\section{Deterministic Algorithms for $\CA$s}\label{sec:algorithms}
	
	In this section, we describe each of the constructive, deterministic algorithms for $\CA$s that were used in our genetic algorithm.
	Each stage starts with every interaction having been covered at least some number of times $\alpha$, and the ``goal'' of the stage is to finish with every interaction covered at least another number of times $\beta$, where $\alpha < \beta$.
	We call $\alpha$ the \textit{current index}, and $\beta$ the \textit{desired index}.
	
	\subsection{Basic}
	
	This stage involves adding one row for each uncovered interaction with index between $\alpha$ and $\beta$. 
	For all entries in the rows that do not involve a specific interaction, we put a fixed value $f$.
	The advantage of this method is clear, in that virtually no additional computation is needed; however, this method suffers by having value $f$ appear in many places; further, interactions that involve many occurrences of $f$ are covered many times, whereas for other interactions, not nearly as much.
	
	For these reasons, we make the basic algorithm more adaptive.
	Instead of inserting the fixed value $f$, we insert into column $c$ any entry that occurs least frequently among rows already constructed.
	By doing this, we are attempting to have columns contain all entries as equally often as possible.
	Because this more adaptive method produced either equal or better results than the original one in practice, and has very small computational difference for cost, we refer to the adaptive method as the basic one from here on.
	
	\subsection{Graph Coloring}
	
	Sarkar and Colbourn \cite{sarkar_colbourn_twostage} define an \textit{incompatibility graph} for a set of interactions as follows.
	The vertices are the interactions that are not covered, and an edge is formed between any two interactions $I_1, I_2$ if in some column shared between $I_1, I_2$, the corresponding values are different.
	The smallest number of colors in a proper coloring of such a graph is the minimum number of rows needed to cover all of these remaining interactions.
	
	We could adapt their strategy for higher-index covering arrays by simply iterating their method multiple times to achieve the desired index.
	However, we can build a single graph and solve the coloring problem in one stage for higher index, as follows.

	The vertices of the graph are all interactions $I$ that are covered at least $\alpha$ times, and fewer than $\beta$ times, paired with an integer $s_I$ with $\alpha \le s_I < \beta$.
	To form edges, let $I_1$ and $I_2$ be two such interactions, with values $s_{I_1}, s_{I_2}$.
	If $I_1, I_2$ share a column with different symbols (regardless of the values of $s_{I_1}, s_{I_2}$), form an edge.
	Otherwise, if $I_1, I_2$ involve the \textit{same} set of columns, but $s_{I_1} \ne s_{I_2}$, we form an edge also.
	This graph has a proper coloring with $N$ colors if and only if $N$ rows can be formed to cover all such interactions.
	
	Since there is no known polynomial-time algorithm for the graph coloring problem, we use the following two ``greedy'' graph coloring algorithms: ``largest first,'' and ``smallest last.''
	For both, the vertices are given an order.
	The former sorts the vertices in non-increasing order of degree.
	The latter sorts the vertices as $v_1, \cdots, v_n$ whenever $v_i$ has the minimum degree in the maximal subgraph among the vertices $v_1, \cdots, v_i$ for every $i$.
	In all cases, the graph coloring algorithm works as follows.
	We process the vertices in the given order according to the method.
	Additionally, an order on the available colors is made; initially, only one color is available.
	For an uncolored vertex $v$, we observe the neighbors of $v$.
	If any color available is possible to be assigned to $v$ such that no neighbor of $v$ has the same color, we choose the ``smallest'' color.
	Otherwise, we allocate a new color, and assign $v$ to that color.
	At the end, the number of colors allocated is the number of corresponding rows in the covering array.
	
	\subsection{Density}\label{subsec:density}
	
	The probabilistic method \cite{alon2004probabilistic} for covering arrays \cite{sarkar2016partial} says that for any $t, k, v$ where $v, t$ are fixed, there must exist some $N = \Theta(\log k)$ for which a covering array on that many rows exists.
	One can show that the optimal number of rows for a higher-index $\CA$ asymptotically is competitively small compared to $O(\lambda \cdot \log k)$, involving juxtaposing the array $\lambda$ times \cite{dougherty2019hash}.
	However, the method does not produce the array, but merely shows that such an array exists.
	
	One can turn this idea into an easy randomized algorithm that generates the array one row at a time, as follows.
	Initially, have $\A$ be an empty array.
	Suppose that $\rho$ interactions remain to be covered.
	The probability that any one of these interactions is covered in a row with entries chosen uniformly at random is $1/v^t$.
	Repeatedly generate rows $R_1, R_2, \cdots$ uniformly at random until some row $R_i$ covers a number of interactions for the first time $c$, where $c \ge \rho / v^t$.
	When such a row is found, append $R_i$ to $\A$, and update the list of interactions not covered.
	An easy analysis shows that when all interactions are covered, $\A$ will have size at most a constant times $\log k$ (where the constant depends on $v, t$).
	
	It is desired to have a deterministic algorithm that has this property, so that a guarantee on the number of rows produced can be made.
	Bryce and Colbourn \cite{bryce2009density} designed the ``conditional expectation'' (or \textit{density}) algorithm, which is deterministic, and a covering array is produced that meets the same logarithmic bound.
	However, their methods do not immediately translate to covering arrays of higher index.
	
	Our method, which generalizes their work, is as follows: let $p = 1/v^t$ be the probability of coverage in a random row as before.
	The probability that an interaction is not $\lambda$-covered in $N$ rows is $\sum_{i=0}^{\lambda-1} {N \choose i} p^i (1-p)^{N-i}$.
	The expected number of interactions not $\lambda$-covered after $N$ rows, if chosen uniformly at random, is ${k \choose t}$ times this probability (or, in general, it is the number of uncovered interactions times this probability).
	When this expectation $E(N)$ is strictly less than 1, a higher-index $\CA$ exists.
	
	We now state our constructive algorithm.
	First, find the smallest $N$ such that $E(N) < 1$.
	Let $A$ be a covering array to-be-built (empty or not).
	Let $T$ be a $t$-way interaction, and $r$ a to-be-completed row that has some column $c$ not fixed to a value.
	For $c$, choose a factor $c_f$.
	Now, determine the expected number of remaining interactions left not $\lambda$-covered \textit{in the remaining $N$-$|A|$ incomplete rows}, if these rows are chosen at random and column $c$ of row $r$ is fixed to $c_f$.
	Then, choose any factor for column $c$ that \textit{minimizes} this expectation.
	Once all entries of row $r$ are fixed, we decrease the choice for $N$ according to how much the expectation decreased after generating $r$. 
	At a high level, $N$ is the estimate on how many rows are needed, and when row is completed, the estimate is updated accordingly (for details, see \cite{dougherty2019hash}).
	
	\section{Experiments}
	
	We designed our experiments to answer the following research questions:
	\begin{itemize}
		\item RQ1: Do multiple stages improve the sizes of $\CA$s of higher-index compared to a single stage?
		\item RQ2: Do multiple stages improve the computational cost in generating such $\CA$s compared to a single stage?
		\item RQ3: For ``optimal'' multiple-stage selections, is there a pattern in regard to algorithms chosen or index values, and how many stages are best for a given parameter situation?
	\end{itemize}
	
	To address these questions, we implemented a genetic algorithm with individuals encoding the multiple stages of algorithms, as well as the corresponding index for each.
	The representation of an individual is a multi-stage selection $\MS\langle A_1(\lambda_1), \cdots, A_m(\lambda_m) \rangle$.
	
	\subsection{Mutation}
	Here we describe the mutation operators used in our genetic algorithm.
	Suppose that the individual is $\MS\langle A_1(\lambda_1), \cdots, A_m(\lambda_m) \rangle$.
	For mutation, there are several operators that we have employed:
	\begin{itemize}
		\item Append: choose an algorithm $\B$ uniformly at random from the ones listed in \Cref{sec:algorithms}.
		Choose any $A_i$ uniformly at random with $\lambda_i > 1$.
		Then, append $\B$ to the individual (at the end) with index $\lambda_{m+1}$ to be a randomly chosen integer between 1 and $\lambda_i-1$ (inclusive).
		We also reduce the index of $A_i$ by $\lambda_{m+1}$.
		If all of the $\lambda_i = 1$, then we remove a randomly chosen $A_i$ before adding $\B$ with its index being 1.
		
		\item Swap: swap the order of two randomly chosen $A_i$ and $A_j$ and their indexes ($i \ne j$).
		\item Index Transfer: take two randomly chosen $A_i$ and $A_j$ ($i \ne j$), subtract a randomly chosen integer from one of their indexes, and add the same integer to the other index.
		\item Modify: substitute a randomly chosen $A_i$ for any other algorithm in its place, with the same index.
		\item Join: randomly choose $A_i$ and $A_j$ ($i \ne j$), remove them from the individual, and re-insert a new algorithm $A_k$, chosen uniformly at random from $\{A_i, A_j\}$; furthermore, the new corresponding index is $\lambda_i + \lambda_j$.
	\end{itemize}
	
	When an individual $I$ is to be mutated, one algorithm from these will be selected uniformly at random to apply to $I$.
	The selection of these operations was to allow for individuals of all types, including the first stage being of high index, many stages with low indexes each, any combination of algorithms including adjacent repetitions, etc. 
	
	\subsection{Crossover}
	
	Crossover works as follows.
	Suppose that the two individuals are of the form $\MS\langle A_1(\lambda_1), \cdots, A_m(\lambda_m) \rangle$, and $\MS\langle B_1(\gamma_1), \cdots, B_n(\gamma_n) \rangle$.
	We pick two positive integers $a, b$ at random such that $a+b \le \lambda$.
	We form a new individual by choosing a random nonempty subset of size $a$ from $\{A_1, \cdots, A_m\}$, another of size $b$ from $\{B_1, \cdots, B_n\}$, and composing them together, ordering the chosen algorithms at random.
	Suppose our formed individual is $\MS\langle C_1(\beta_1), \cdots, C_{a+b}(\beta_{a+b}) \rangle$, where the $\beta_i$ come from the corresponding individual and algorithm.
	
	Note that the sum of the $\beta_i$ in our new individual may exceed $\lambda$.
	If this is the case, we choose a $\beta_i > 1$ at random and decrease $\beta_i$ by 1; repeat this procedure until the sum of the $\beta_i$ is equal to $\lambda$.
	If the sum originally is less than $\lambda$, then we perform nearly the opposite operation: choose any $\beta_i$ (may be equal to 1) at random, and increase $\beta_i$ by 1; repeat until the sum is $\lambda$.
	
	The fitness of an individual, naturally, is a tuple $(N, T)$, where $N$ is the number of rows in the produced covering array, and $T$ is the total computational time.
	When we sort the population by fitness, we sort by $N$ first, and then if two individuals have the same number of rows, we sort them by their $T$ values.
	We used the NSGA-II algorithm \cite{deb2002fast} to observe the Pareto front of the population.
	
	\subsection{Results}
	
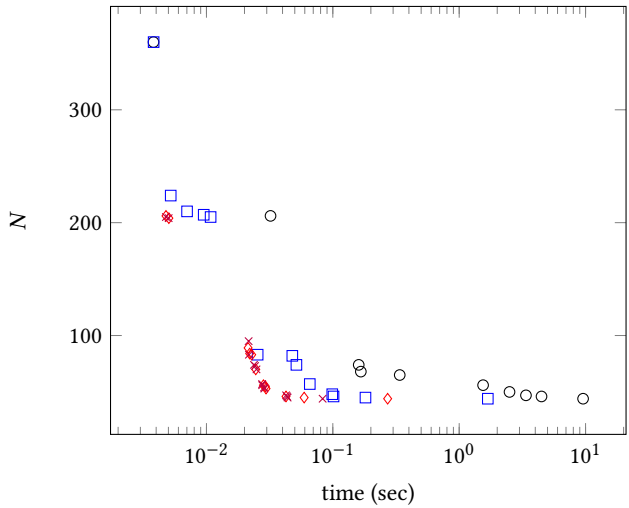
\begin{figure}
		\begin{tikzpicture}
		\begin{axis}[%
		xmode=log,
		xlabel={time (sec)},
		ylabel={$N$},
		scatter/classes={%
			a={mark=o,draw=black},
			b={mark=square,draw=blue},
			c={mark=diamond,draw=red},
			d={mark=x,draw=purple}
		}]
		\addplot[scatter,only marks,%
		scatter src=explicit symbolic]%
		table[meta=label] {
			x y label
			9.55871319770813 44 a
			4.481946229934692 46 a
			3.3759171962738037 47 a
			2.5019760131835938 50 a
			1.5457179546356201 56 a
			0.3384430408477783 65 a
			0.16659808158874512 68 a
			0.16030073165893555 74 a
			0.03214883804321289 206 a
			0.0038101673126220703 360 a
			1.6848101615905762 44 b
			0.18130159378051758 45 b
			0.10130667686462402 46 b
			0.0989835262298584 48 b
			0.06574130058288574 57 b
			0.05150771141052246 74 b
			0.047799110412597656 82 b
			0.025421142578125 83 b
			0.010786771774291992 205 b
			0.009514570236206055 207 b
			0.00702357292175293 210 b
			0.005214691162109375 224 b
			0.0038199424743652344 360 b
			0.2715456485748291 44 c
			0.059326887130737305 45 c
			0.04260754585266113 46 c
			0.029613018035888672 53 c
			0.02926325798034668 55 c
			0.028258085250854492 56 c
			0.024564027786254883 70 c
			0.022802352905273438 83 c
			0.021982908248901367 84 c
			0.02135610580444336 89 c
			0.005042552947998047 204 c
			0.0047838687896728516 206 c
			0.08304762840270996 44 d
			0.043886661529541016 45 d
			0.04340672492980957 46 d
			0.04268074035644531 47 d
			0.028701305389404297 53 d
			0.028608083724975586 54 d
			0.027607202529907227 56 d
			0.027590274810791016 57 d
			0.024891138076782227 70 d
			0.0242464542388916 73 d
			0.023774385452270508 74 d
			0.021682262420654297 83 d
			0.021566152572631836 95 d
			0.005022764205932617 204 d
			0.004779815673828125 205 d
		};
		\end{axis}
		
		\end{tikzpicture}
		\caption{Pareto Fronts of $\CA_\lambda(N;t,k,v)$s formed, where $\lambda=10, t=2, k=10, v=2$, for Generations 1, 10, 50, and 100.
			Circles are Generation 1, Squares are Generation 10, Diamonds are Generation 50, and Crosses are Generation 100.
			\label{fig:example_pareto_front}}
	\end{figure}

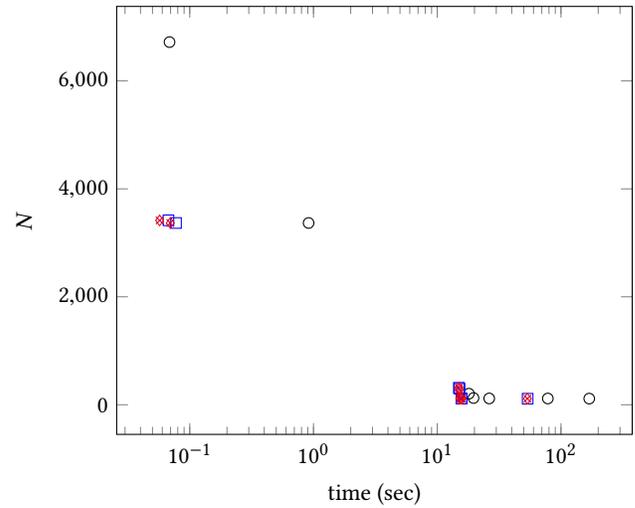
\begin{figure}
		\begin{tikzpicture}
		\begin{axis}[%
		xmode=log,
		xlabel={time (sec)},
		ylabel={$N$},
		scatter/classes={%
			a={mark=o,draw=black},
			b={mark=square,draw=blue},
			c={mark=diamond,draw=red},
			d={mark=x,draw=purple}
		}]
		\addplot[scatter,only marks,%
		scatter src=explicit symbolic]%
		table[meta=label] {
			x y label
			169.25479006767273 114 a
			78.3174250125885 116 a
			26.308947801589966 117 a
			19.643487215042114 125 a
			18.03126287460327 205 a
			0.9132859706878662 3368 a
			0.06864023208618164 6720 a
			53.795217990875244 112 b
			15.757892370223999 114 b
			15.698622226715088 115 b
			15.151724338531494 296 b
			14.96637225151062 315 b
			0.0774393081665039 3368 b
			0.06730341911315918 3418 b
			53.55049395561218 112 c
			15.802470922470093 113 c
			15.744463443756104 114 c
			15.286712408065796 115 c
			15.205374240875244 118 c
			15.055338859558105 276 c
			14.962716102600098 315 c
			0.06963419914245605 3368 c
			0.05710935592651367 3416 c
			53.55049395561218 112 d
			15.802470922470093 113 d
			15.744463443756104 114 d
			15.286712408065796 115 d
			15.205374240875244 118 d
			15.055338859558105 276 d
			14.962716102600098 315 d
			0.06963419914245605 3368 d
			0.05710935592651367 3416 d
		};
		\end{axis}
		
		\end{tikzpicture}
		\caption{Pareto Front of $\CA_\lambda(N;t,k,v)$s formed, where $\lambda=5, t=4, k=10, v=2$.
			Circles are Generation 1, Squares are Generation 10, Diamonds are Generation 50, and Crosses are Generation 100.
			\label{fig:example_pareto_front_3}}
	\end{figure}

	We evaluated our genetic algorithm on various parameter situations for $t \in \{2, 3, 4\}$.
	The results are reported in \Cref{tbl:ga_results}.
	Here is a short-hand list of all the algorithms reported there:
	\begin{itemize}
		\item Basic: $\B$
		\item Coloring (largest first): $\Largest$
		\item Coloring (smallest last): $\Smallest$
		\item Density: $\D$
	\end{itemize}
	For ease of presentation in the table, we shorten $\MS\langle A_1(\lambda_1), \cdots, A_m(\lambda_m)\rangle$ to be a list of the algorithm names, with a subscript indicating the index.
	For example, the individual representing $\MS\langle\B(1), \Smallest(1), \B(1), \Largest(2)\rangle$ is instead written as $\B_1, \Smallest_1, \B_1, \Largest_2$.
	In all our experiments, our population had 300 individuals, and we report results for up to 100 generations.
	\Cref{tbl:single_alg_results} contains the results for a single stage.
	Further, in \Cref{fig:example_pareto_front,fig:example_pareto_front_2,fig:example_pareto_front_3}, we provide figures of scatter plots for the situations:
	\begin{itemize}
		\item $t=2, k=10, v=2, \lambda=10$;
		\item $t=3, k=10, v=2, \lambda=5$; and
		\item $t=4, k=10, v=2, \lambda=5$.
	\end{itemize} 
	The horizontal axis (time) is given in logarithmic scale.
	The circles indicate generation 1, squares are generation 10, diamonds are generation 50, and crosses are generation 100.
	These figures show the Pareto fronts of the genetic algorithm for each of the same generations examined, and are representative over all parameter situations given in \Cref{tbl:ga_results}.

	\begin{table*}[h]
		
		\begin{tabular}{|l|l|l|l|l|l|l|l|l|l|l|l|}
			\hline
			\multirow{2}{*}{$t$} & \multirow{2}{*}{$k$} & \multirow{2}{*}{$v$} & \multirow{2}{*}{$\lambda$} & \multicolumn{2}{c|}{Generation 1}                                                                                                                                                & \multicolumn{2}{c|}{Generation 10}                                                                                                                                                                                       & \multicolumn{2}{c|}{Generation 50}                                                                                                                                                                                                                                                                                            & \multicolumn{2}{c|}{Generation 100}                                                                                                                                                                                                                                                                                                                                  \\ \cline{5-12} 
			&                      &                      &                            & Lowest $N$                                                                             & Lowest Time                                                                             & Lowest $N$                                                                                                          & Lowest Time                                                                                        & Lowest $N$                                                                                                                               & Lowest Time                                                                                                                                                                        & Lowest $N$                                                                                                                                                                      & Lowest Time                                                                                                                                                                        \\ \hline
			2                    & 10                   & 2                    & 5                          & \begin{tabular}[c]{@{}l@{}}24, 0.482\\ $\Smallest_4, \Largest_1$\end{tabular}          & \begin{tabular}[c]{@{}l@{}}189, 0.003\\ $\B_1,\Smallest_1,\B_1,\Largest_2$\end{tabular} & \begin{tabular}[c]{@{}l@{}}24, 0.157\\ $\Smallest_2, \Smallest_1, \D_1, \Largest_1$\end{tabular}                    & \begin{tabular}[c]{@{}l@{}}189, 0.003\\ $\B_1, \B_2, \Largest_2$\end{tabular}                      & \begin{tabular}[c]{@{}l@{}}24. 0.071\\ $\Smallest_1,\D_1,\Largest_1$,\\ $\Largest_1,\Smallest_1$\end{tabular}                            & \begin{tabular}[c]{@{}l@{}}189, 0.003\\ $\B_1,\Largest_1$,\\ $\Largest_1,\Largest_2$\end{tabular}                                                                                  & \begin{tabular}[c]{@{}l@{}}24. 0.071\\ $\Smallest_1,\D_1,\Largest_1$,\\ $\Largest_1,\Smallest_1$\end{tabular}                                                                   & \begin{tabular}[c]{@{}l@{}}189, 0.003\\ $\B_1,\Largest_1$,\\ $\Largest_1,\Largest_2$\end{tabular}                                                                                  \\ \hline
			2                    & 10                   & 2                    & 10                         & \begin{tabular}[c]{@{}l@{}}44, 9.559\\ $\D_9, \Smallest_1$\end{tabular}                & \begin{tabular}[c]{@{}l@{}}360, 0.004\\ $\B_2,\Largest_2,\B_6$\end{tabular}             & \begin{tabular}[c]{@{}l@{}}44, 1.685\\ $\D_1, \Smallest_1, \D_1$,\\ $\D_3, \D_4$\end{tabular}                       & \begin{tabular}[c]{@{}l@{}}360, 0.004\\ $\B_2, \B_8$\end{tabular}                                  & \begin{tabular}[c]{@{}l@{}}44, 0.272\\ $\Smallest_1, \Largest_1,\Smallest_1$,\\ $\D_1,\Largest_1,\D_1$,\\ $\D_2,\Largest_2$\end{tabular} & \begin{tabular}[c]{@{}l@{}}206, 0.004\\ $\B_1,\Largest_1,\Smallest_1$,\\ $\Smallest_1,\Smallest_1,\Smallest_1$,\\ $\Smallest_1,\Largest_1,\Largest_1$,\\ $\Largest_1$\end{tabular} & \begin{tabular}[c]{@{}l@{}}44, 0.083\\ $\Smallest_1,\Smallest_1,\D_1$,\\ $\Largest_1,\Largest_1,\Largest_1$,\\ $\Largest_1,\Smallest_1,\Largest_1$,\\ $\Largest_1$\end{tabular} & \begin{tabular}[c]{@{}l@{}}206, 0.004\\ $\B_1,\Largest_1,\Smallest_1$,\\ $\Smallest_1,\Smallest_1,\Smallest_1$,\\ $\Smallest_1,\Largest_1,\Largest_1$,\\ $\Largest_1$\end{tabular} \\ \hline
			2                    & 10                   & 3                    & 5                          & \begin{tabular}[c]{@{}l@{}}56, 27.628\\ $\D_5$\end{tabular}                            & \begin{tabular}[c]{@{}l@{}}816, 0.009\\ $\B_2,\Largest_1,\Largest_2$\end{tabular}       & \begin{tabular}[c]{@{}l@{}}55, 4.838\\ $\Smallest_1,\D_1,\D_3$\end{tabular}                                         & \begin{tabular}[c]{@{}l@{}}442, 0.007\\ $\B_1,\Largest_1,\B_1$,\\ $\Largest_1,\B_1$\end{tabular}   & \begin{tabular}[c]{@{}l@{}}55, 4.838\\ $\Smallest_1,\D_1,\D_3$\end{tabular}                                                              & \begin{tabular}[c]{@{}l@{}}429, 0.007\\ $\B_1,\B_1,\B_1$,\\ $\B_1,\Largest_1$\end{tabular}                                                                                         & \begin{tabular}[c]{@{}l@{}}55, 4.838\\ $\Smallest_1,\D_1,\D_3$\end{tabular}                                                                                                     & \begin{tabular}[c]{@{}l@{}}429, 0.007\\ $\B_1,\B_1,\B_1$,\\ $\B_1,\Largest_1$\end{tabular}                                                                                         \\ \hline
			2                    & 20                   & 2                    & 5                          & \begin{tabular}[c]{@{}l@{}}28, 2.179\\ $\Smallest_1, \Smallest_3$\end{tabular}         & \begin{tabular}[c]{@{}l@{}}770, 0.019\\ $\B_1,\Smallest_4$\end{tabular}                 & \begin{tabular}[c]{@{}l@{}}27, 0.369\\ $\Smallest_1,\Smallest_1,\Largest_1$,\\ $\Largest_1,\Largest_1$\end{tabular} & \begin{tabular}[c]{@{}l@{}}769, 0.017\\ $\B_1,\Largest_1,\Largest_3$\end{tabular}                  & \begin{tabular}[c]{@{}l@{}}27, 0.369\\ $\Smallest_1,\Smallest_1,\Largest_1$,\\ $\Largest_1,\Largest_1$\end{tabular}                      & \begin{tabular}[c]{@{}l@{}}769, 0.017\\ $\B_1,\Largest_1,\Largest_3$\end{tabular}                                                                                                  & \begin{tabular}[c]{@{}l@{}}27, 0.369\\ $\Smallest_1,\Smallest_1,\Largest_1$,\\ $\Largest_1,\Largest_1$\end{tabular}                                                             & \begin{tabular}[c]{@{}l@{}}769, 0.017\\ $\B_1,\Largest_1,\Largest_3$\end{tabular}                                                                                                  \\ \hline
			2                    & 20                   & 2                    & 10                         & \begin{tabular}[c]{@{}l@{}}48, 26.623\\ $\D_1,\D_1$,\\ $\Smallest_1,\D_7$\end{tabular} & \begin{tabular}[c]{@{}l@{}}1520, 0.042\\ $\B_2,\D_1,\Largest_7$\end{tabular}            & \begin{tabular}[c]{@{}l@{}}48, 12.732\\ $\D_1,\Smallest_1,\Largest_1$,\\ $\D_1,\D_1,\D_5$\end{tabular}              & \begin{tabular}[c]{@{}l@{}}787, 0.024\\ $\B_1,\Largest_2$,\\ $\Largest_6,\Smallest_1$\end{tabular} & \begin{tabular}[c]{@{}l@{}}48, 7.772\\ $\D_1,\Smallest_1,\Largest_1$,\\ $\D_1,\D_1,\D_4,\Smallest_1$\end{tabular}                        & \begin{tabular}[c]{@{}l@{}}787, 0.024\\ $\B_1,\Largest_2$,\\ $\Largest_6,\Smallest_1$\end{tabular}                                                                                 & \begin{tabular}[c]{@{}l@{}}48, 7.772\\ $\D_1,\Smallest_1,\Largest_1$,\\ $\D_1,\D_1,\D_4,\Smallest_1$\end{tabular}                                                               & \begin{tabular}[c]{@{}l@{}}787, 0.024\\ $\B_1,\Largest_2$,\\ $\Largest_6,\Smallest_1$\end{tabular}                                                                                 \\ \hline
			3                    & 10                   & 2                    & 5                          & \begin{tabular}[c]{@{}l@{}}48, 33.537\\ $\D_5$\end{tabular}                            & \begin{tabular}[c]{@{}l@{}}1050, 0.023\\ $\B_1,\Smallest_1,\B_3$\end{tabular}           & \begin{tabular}[c]{@{}l@{}}48, 3.381\\ $\D_1,\D_1,\D_1$,\\ $\Largest_1,\D_1$\end{tabular}                           & \begin{tabular}[c]{@{}l@{}}976, 0.015\\ $\B_1,\B_1,\B_1$,\\ $\Largest_1,\B_1$\end{tabular}         & \begin{tabular}[c]{@{}l@{}}48, 3.381\\ $\D_1,\D_1,\D_1$,\\ $\Largest_1,\D_1$\end{tabular}                                                & \begin{tabular}[c]{@{}l@{}}976, 0.015\\ $\B_1,\B_1,\B_1$,\\ $\Largest_1,\B_1$\end{tabular}                                                                                         & \begin{tabular}[c]{@{}l@{}}48, 3.381\\ $\D_1,\D_1,\D_1$,\\ $\Largest_1,\D_1$\end{tabular}                                                                                       & \begin{tabular}[c]{@{}l@{}}976, 0.015\\ $\B_1,\B_1,\B_1$,\\ $\Largest_1,\B_1$\end{tabular}                                                                                         \\ \hline
			4                    & 10                   & 2                    & 5                          & \begin{tabular}[c]{@{}l@{}}114, 169.255\\ $\D_1,\D_4$\end{tabular}                     & \begin{tabular}[c]{@{}l@{}}6720, 0.093\\ $\B_2, \B_3$\end{tabular}                      & \begin{tabular}[c]{@{}l@{}}112, 53.795\\ $\Smallest_1,\Largest_1,\D_3$\end{tabular}                                 & \begin{tabular}[c]{@{}l@{}}3418, 0.067\\ $\B_1,\Smallest_1$,\\ $\B_1,\Smallest_2$\end{tabular}     & \begin{tabular}[c]{@{}l@{}}112, 53.795\\ $\Smallest_1,\Largest_1,\D_3$\end{tabular}                                                      & \begin{tabular}[c]{@{}l@{}}3416, 0.070\\ $\B_1,\B_1,\Largest_1$,\\ $\Largest_1,\Largest_1$\end{tabular}                                                                            & \begin{tabular}[c]{@{}l@{}}112, 53.795\\ $\Smallest_1,\Largest_1,\D_3$\end{tabular}                                                                                             & \begin{tabular}[c]{@{}l@{}}3416, 0.070\\ $\B_1,\B_1,\Largest_1$,\\ $\Largest_1,\Largest_1$\end{tabular}                                                                            \\ \hline
		\end{tabular}
		\caption{Genetic Algorithm Results, with two individuals reported for each parameter situation: the one with the lowest $N$, and the one with the lowest time, at generations 1, 10, 50, and 100.
			The two numbers in each entry signify that individual's $N$ and time, as well as the algorithms with indexes chosen as subscripts.\label{tbl:ga_results}}
	\end{table*}
	
	\begin{table}[]
		\begin{tabular}{|l|l|l|l|l|l|}
			\hline
			$t$ & $k$ & $v$ & $\lambda$ & Lowest $N$            & Lowest Time            \\ \hline
			2   & 10  & 2   & 5         & 26, 2.976, $\D_{5}$   & 900, 0.036, $\B_{5}$   \\ \hline
			2   & 10  & 2   & 10        & 44, 10.855, $\D_{10}$ & 1800, 0.069, $\B_{10}$ \\ \hline
			2   & 10  & 3   & 5         & 56, 27.120, $\D_{5}$  & 2025, 0.083, $\B_{5}$  \\ \hline
			2   & 20  & 2   & 5         & 31, 23.994, $\D_5$    & 3800, 0.562, $\B_5$    \\ \hline
			2   & 20  & 2   & 10        & 49, 97.389, $\D_{10}$ & 7600, 1.107, $\B_{10}$ \\ \hline
			3   & 10  & 2   & 5         & 48, 34.623, $\D_5$    & 4800, 0.473, $\B_5$    \\ \hline
			4   & 10  & 2   & 5         & 121, 262.64, $\D_5$ &  16800, 3.077, $\B_5$     \\ \hline
		\end{tabular}
		\caption{Individuals composed of a single stage with the smallest number of rows $N$ and smallest time $T$ for the same parameter situations as in \Cref{tbl:ga_results}.\label{tbl:single_alg_results}}
	\end{table}
	
	\begin{figure}
		\begin{tikzpicture}
		\begin{axis}[%
		xmode=log,
		xlabel={time (sec)},
		ylabel={$N$},
		scatter/classes={%
			a={mark=o,draw=black},
			b={mark=square,draw=blue},
			c={mark=diamond,draw=red},
			d={mark=x,draw=purple}
		}]
		\addplot[scatter,only marks,%
		scatter src=explicit symbolic]%
		table[meta=label] {
			x y label
			33.537331342697144 48 a
			15.45622992515564 50 a
			14.17971134185791 51 a
			9.028223752975464 56 a
			4.8995349407196045 57 a
			2.81516695022583 61 a
			1.4164671897888184 80 a
			0.24357843399047852 968 a
			0.22501897811889648 982 a
			0.02308058738708496 1050 a
			3.3805758953094482 48 b
			2.7771787643432617 50 b
			2.4818179607391357 52 b
			2.3427417278289795 53 b
			1.988403081893921 55 b
			1.8816900253295898 56 b
			1.801419734954834 57 b
			1.799726963043213 58 b
			1.7611360549926758 59 b
			1.463813066482544 71 b
			1.3777692317962646 72 b
			1.130511999130249 75 b
			1.0986413955688477 122 b
			0.04953145980834961 968 b
			0.015101432800292969 976 b
			3.3369197845458984 48 c
			2.75403094291687 50 c
			2.393850088119507 52 c
			2.2292380332946777 53 c
			2.146951675415039 54 c
			1.7893469333648682 55 c
			1.6592941284179688 56 c
			1.6573240756988525 57 c
			1.3993520736694336 70 c
			1.3334622383117676 71 c
			1.0007522106170654 72 c
			0.9843671321868896 80 c
			0.9795262813568115 81 c
			0.9783902168273926 138 c
			0.9455051422119141 268 c
			0.8970022201538086 314 c
			0.04727458953857422 968 c
			3.390645980834961 48 d
			2.776526927947998 50 d
			2.402292013168335 52 d
			2.2373909950256348 53 d
			2.153233051300049 54 d
			1.7984521389007568 55 d
			1.6610047817230225 56 d
			1.4266631603240967 70 d
			1.3550500869750977 71 d
			1.009904146194458 72 d
			1.0021162033081055 75 d
			0.9878308773040771 80 d
			0.9845459461212158 81 d
			0.9796762466430664 138 d
			0.949242115020752 268 d
			0.8892562389373779 314 d
			0.048476457595825195 968 d
		};
		\end{axis}
		
		\end{tikzpicture}
		\caption{Pareto Front of $\CA_\lambda(N;t,k,v)$s formed, where $\lambda=5, t=3, k=10, v=2$.
			Circles are Generation 1, Squares are Generation 10, Diamonds are Generation 50, and Crosses are Generation 100.
			\label{fig:example_pareto_front_2} }
	\end{figure}
	
	\section{Discussion}\label{sec:discussion}
	
	We discuss each of the research questions in turn.
	For RQ1, it is quite evident that multiple stages do improve the sizes of covering arrays. 
	For example, observe the situation of $t=4, k=10, v=2, \lambda=5$: the most fit individual produced had 112 rows, whereas even a two-stage solution had 114 for its fittest individual.
	Even though the individuals with smallest $N$ did not improve the number of rows significantly, what is more striking is when one observes \Cref{fig:example_pareto_front,fig:example_pareto_front_2,fig:example_pareto_front_3}, especially \Cref{fig:example_pareto_front}: the distribution across the vertical axis decreases substantially as the number of generations increases.
	
	For RQ2, it is even more evident that multiple stages improve the computational time to construct covering arrays.
	A striking example is when $t=2, k=10, v=2, \lambda=10$; the individual with 44 rows in Generation 1 had a run-time of 9.559 seconds, and by Generation 100, an individual exists with the same number of rows but only takes 0.083 seconds, a improvement by over two orders of magnitude. 
	In fact, the construction time difference between this final individual and the one with 206 rows is so small that the latter can be completely ignored.
	Further proof of the strong improvement is in \Cref{fig:example_pareto_front,fig:example_pareto_front_2}, in that the Pareto fronts shift left considerably as the number of generations increases.
	
	For RQ3, we note that for all entries in the ``Lowest $N$'' columns, the first algorithm chosen is either density ($\D$) or ``smallest last'' ($\Smallest$). 
	Also, in the ``Lowest Time'' columns, the first algorithm chosen is always ``Basic'' ($\B$).
	Interestingly enough, ``Largest First'' ($\Largest$) never was the first algorithm.
	An analysis of the output data reveals that $\Largest$ does in fact produce covering arrays with a small $N$ relative to the entire distribution, but it is not the smallest $N$ found.
	An explanation of this may be that this algorithm has to work with a set of interactions that has many symmetries, and one choice among many vertices with high degree may end up with more rows than another choice does.
	
	Furthermore, at Generation 10 and after (apart from two exceptions), the index for the first algorithm is always 1.
	The index for most of the later stages is 1, or small relative to the index, apart from a few exceptions.
	This confirms our intuition, in that a low-index first stage is computationally efficient, produces a small number of rows, and simultaneously covers many interactions.
	
	Note that the genetic algorithm has very much the same effect no matter if $t$ increases, $k$ increases, $v$ increases, or $\lambda$ increases.
	However, the extent to which the computational time decreases appears to be smaller as $t$ increases, such as for the $t=4$ situation, where the timeof the ``Lowest $N$'' individual only decreases by 68.2\%.
	
	An interesting point to discuss relates not to the genetic algorithm, but rather to the experiment with a single stage, in \Cref{tbl:single_alg_results}. 
	Observe the last row, with parameters $t =4, k=10, v=2, \lambda=5$: the ``Lowest $N$'' individual had 121 rows, completed in 262.640 seconds, and the selection was $\MS\langle \D(5) \rangle$.
	The density algorithm is computationally intensive, but here we actually found that this algorithm did not take the longest time.
	In fact, $\MS\langle\Largest(5)\rangle$ produced 740 rows in over 430 seconds, and $\MS\langle\Smallest(5)\rangle$ produced 137 rows in over 1000 seconds!
	An explanation is that as the strength $t$ increases, then the graphs being constructed for $\Largest$ and $\Smallest$ are very large. 
	Since there is only one stage, the entire graph on ${k \choose t}v^t$ vertices is created, one for each interaction.
	Note that $\D$ also maintains a structure for all of these interactions.
	But the higher-index formulation of the edges imply that the graph is very highly connected,
	whereas $\D$ only has to maintain a constant times the number of interactions.
	Much of the computation for $\Largest$ and $\Smallest$ was dedicated to managing the (large) graph all at once, since the size of the graph was dominated by the number of edges, not vertices.
	
	\subsection{Genetic Operators Discussion}
	
	It is clear from the results in \Cref{tbl:ga_results} that the two most important operations towards achieving a smaller covering array faster are Append and Modify, since most of the most fit individuals found have many stages with small indexes.
	
	There appears to be a correlation between these two operators, for the sole reason that there are many instances of the same algorithm with the same index being repeated.
	For example, for $t=2, k=10, v=2, \lambda=10$, in Generation 50, the ``Lowest Time'' individual has \textit{six} occurrences of $\Smallest$ with index 1 repeated.
	It would be far simpler to use one instance of $\Smallest$ with index 6.
	
	We explain why this is the case for each of the algorithms in turn.
	For $\B$, there is no heuristic being calculated as rows are generated, since each interaction currently uncovered is put into its own row(s).
	Because there is no additional calculation occurring, $\B$ is fast but produces far too many rows.
	By having multiple adjacent occurrences of $\B$ with index 1, there is a small ``heuristic'' being calculated, which happens between the stages in determining which interactions are left. 
	Since this cost is far less after the first stage is complete, we can now see why this occurs for $\B$.
	
	The explanation for $\Largest$ and $\Smallest$ is nearly identical.
	The advantage of coloring algorithms occurs most when the graph is ``sparsely'' connected; if there are only a few edges comparatively, then the choices made for each of the algorithms will produce a coloring that is closer to the optimal number.
	Suppose we are observing a coloring stage with a relatively large index, and suppose that our current index is $\alpha$, with the target index of $\beta$.
	Then a \textit{clique} (i.e., a set of vertices with every pair connected via an edge) is formed among \textit{all} interactions with the same column set and value set, but different index.
	Therefore, when the index of a stage increases, the graph becomes much more connected than with one stage.
	Like the case with $\B$ before, a stage with high target index has no heuristic other than the choice of vertices, but by having multiple repeated stages with index 1 of the same algorithm, there is a ``heuristic'' created between the stages.
	
	For $\D$, the algorithm is greedy in that it chooses the ``best'' symbol in each position of a row being generated.
	As is shown in \cite{bryce2007one}, the rate of coverage decreases as the number of generated rows increase.
	If we have a stage for density with high target index, then this phenomenon certainly occurs: we saw this in \Cref{fig:two_example_cas}.
	By having multiple stages of $\D$ with index 1, and recalculating the interactions left to cover after a stage is complete, we can now see why multiple sequential stages of $\D$ improve over a single stage with high index.
	
	The crossover operator did not prove to be nearly as powerful as the mentioned mutation operators.
	We can confirm this by a similar experiment that we performed that was mutation-only (i.e., no crossover), and it eventually produced individuals that were either the same, or nearly equivalent, to those shown in \Cref{tbl:ga_results}.
	However, we had to extend the algorithm to nearly 200 generations before these individuals were found.
	Therefore, we can conclude that the crossover operator was helpful only in improving the number of generations in the genetic algorithm to find ``very fit'' individuals.
	
	\section{Conclusions and Future Work}
	
	In this paper, we developed a genetic algorithm that constructs covering arrays of higher index using a sequence of deterministic algorithms.
	This algorithm is a generalization of existing methods but simultaneously addresses the question of redundancy in interaction testing, which has not been examined in any publication as far as we are aware.
	We believe that an exploration of redundancy with covering arrays and similar objects will lead to improved testing practices in systems that are inherently not deterministic.
	
	As a result of our experiments, we can conclude the following different avenues for the construction of covering arrays of higher index.
	A very good solution can be obtained with all indexes set to 1, with the first algorithm dependent on the desired goal.
	\begin{itemize}
		\item If the goal is mostly computational efficiency and very little about $N$, have the first algorithm be $\B$, followed by any choice of greedy graph coloring algorithms.
		Any index can be chosen here, but a lower $N$ is reached with very little additional time if the first index is 1.
		\item If the goal is a balance between computational efficiency and $N$, then have most algorithms be greedy graph coloring ones, either $\Largest$ or $\Smallest$.
		Whichever algorithms are chosen, have $\B$ not be in any stage.
		\item If the goal is to minimize $N$, choose either $\Smallest$ or $\D$ as the first stage, as well as several choices of $\D$ in subsequent stages.
		Whichever algorithms are chosen, have $\B$ not be in any stage.
	\end{itemize}

	Our future work involves introducing randomness into the framework.
	Although random methods do not have guarantees about the size of the produced array, they are often much faster, and the number of rows in practice is often close or better than deterministic algorithms.
	We plan to investigate incorporating randomness into the fitness function while also keeping the simulation time low, which was a distinct advantage in our approach of only having deterministic algorithms.
	One direction that we plan to investigate is to have the framework start with a randomly generated array, have it remain fixed, and then find a sequence of stages starting from this array (our algorithm developed here starts without any rows built).
	
	One aspect of our algorithm worth observing is that it is absolute, in that if a number of rows $N$ and time $T$ are observed for an individual $I$ such that $T$ is the minimum time in the distribution, then it is reported in \Cref{tbl:ga_results}. 
	However \Cref{fig:example_pareto_front,fig:example_pareto_front_2,fig:example_pareto_front_3} show examples of two points $(N_1, T_1)$ and $(N_2, T_2)$ with $T_1 < T_2$, but $N_1$ is much greater than $N_2$. 
	In other words, the time decrease is very small, but the number of rows increase is very large.
	It would be worth exploring an alteration to the Pareto front calculation that picks the point with minimum $N$ and ``small'' $T$ such that the time distance between it and the global minimum time is within a small $\varepsilon$ factor.
	
	Another future work item considers exact methods, such as satisfiability and constraint programming.
	While they guarantee to produce the smallest number of rows possible in a given stage and are deterministic, they are far slower than any of the algorithms we have tested.
	Furthermore, since most of the individuals we have found here have each stage representing a small index, we believe that each algorithm (other than $\B$) produced an array very close to the optimal size for a given stage.
	It would be interesting, however, to see if such exact methods would be useful at a very late stage.
	This is mainly because most deterministic methods often cover many interactions early, leaving a small number of them remaining, and often having trouble covering them in a small number of rows.
	Further, since the number of interactions left is small, the time needed to run these exact methods is far reduced.

\bibliographystyle{ACM-Reference-Format}
\bibliography{sample-bibliography} 

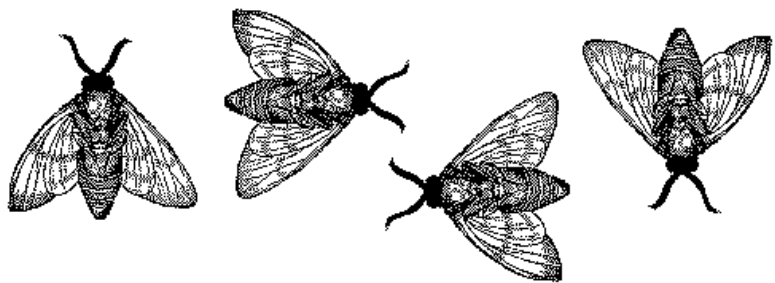
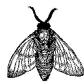
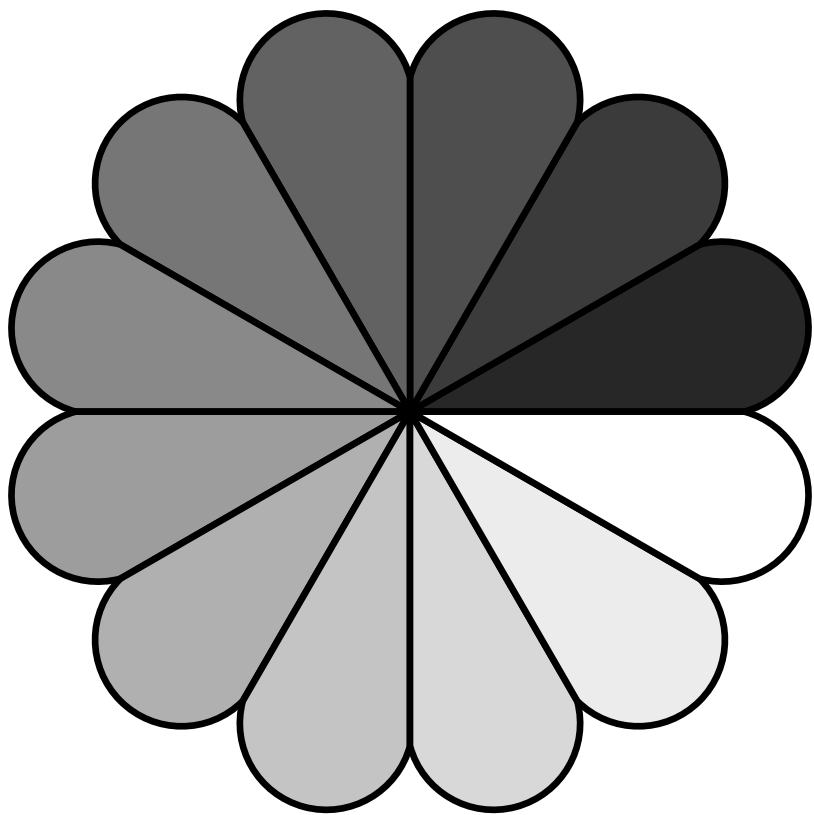
\end{document}